\documentclass[11pt,a4paper]{article}
\usepackage[hyperref]{naaclhlt2019}
\usepackage{times}
\usepackage{latexsym}

\usepackage{url}
\usepackage[pdftex]{graphicx}
\usepackage{amsmath}

\DeclareMathOperator*{\argmin}{arg\,min}
\usepackage{dsfont}
\usepackage{multirow}
\usepackage{todonotes}
\usepackage[labelfont=bf]{caption}
\usepackage{array}

\usepackage[compact]{titlesec}

\aclfinalcopy 
\def\aclpaperid{4092} 

\title{Locale-agnostic Universal Domain Classification Model in Spoken Language Understanding}

\author{Jihwan Lee \\
  Amazon Alexa AI\\
  {\tt jihwl@amazon.com} \\\And
  Ruhi Sarikaya \\
  Amazon Alexa AI\\
  {\tt rsarikay@amazon.com} \\\And
  Young-Bum Kim \\
  Amazon Alexa AI\\
  {\tt youngbum@amazon.com}}

\date{}

\begin{document}
\maketitle
\begin{abstract}
  In this paper, we introduce an approach for leveraging available data across multiple locales sharing the same language to 1) improve domain classification model accuracy in Spoken Language Understanding and user experience even if new locales do not have sufficient data and 2) reduce the cost of scaling the domain classifier to a large number of locales. We propose a locale-agnostic universal domain classification model based on selective multi-task learning that learns a joint representation of an utterance over locales with different sets of domains and allows locales to share knowledge selectively depending on the domains. The experimental results demonstrate the effectiveness of our approach on domain classification task in the scenario of multiple locales with imbalanced data and disparate domain sets. The proposed approach outperforms other baselines models especially when classifying locale-specific domains and also low-resourced domains.
\end{abstract}

\section{Introduction}
\label{label:introduction}
Recent success of intelligent personal digital assistants (IPDA) such as Amazon Alexa, Google Assistant, Apple Siri, Microsoft Cortana~\citep{sarikaya2017technology, sarikaya2016overview} in USA has led to their expansion to multiple locales and languages. Some of those virtual assistant systems have been released in the United States (US), the United Kingdom (GB), Canada (CA), India (IN), and so on.
Such expansion typically leads to building a separate domain classification model for each new locale, and it brings two challenging issues: 1) having a separate model per locale becomes a bottleneck for rapid scaling of virtual assistant due to the resource and maintenance costs that grow linearly with the number of locales, and 2) new locales typically comes without much training data and cannot take full advantage of useful data available in other mature locales to achieve the high model accuracy.


In this study, we propose a new approach that reduces the cost of scaling natural language understanding to a large number of locales, given the sufficient amount of data in one of the locales of that language, while achieving high domain classification accuracy over all locales. The approach is based on a multi-task learning framework that aims to share available data to learn a joint representation, and we introduce a way to selectively share knowledge across locales while considering locale-specificity in the joint learning. Multi-task learning has been widely used to tackle the problem of low-resource tasks or leveraging data between correlated targets~\citep{Liu2017AdversarialML,ruder2018strong,augenstein2018multi,peters2017semi,kim2017onenet}, but none of them consider locale-specificity when sharing knowledge to learn a joint representation.


We evaluate our proposed approach on the real-world utterance data spoken by customers to an intelligent personal digital assistant across different locales. The experimental results empirically demonstrate that the proposed universal model scales to multiple locales, while achieving higher domain classification accuracy compared to competing locale-unified models as well as per-locale separate models. The proposed model named universal model is able to successfully predict domains for locale-specific utterances while sharing common knowledge across locales without sacrificing the accuracy of predicting locale-independent domains.

The paper is organized as follows. In Section~\ref{sec:motivation}, we discuss several design considerations that motivate our model design. In Section~\ref{label:universal_model}, we define the problem of domain classification with multiple locales that have different domain sets, and then introduce a novel universal domain classification model with several technical details. We present our experimental observations over different approaches on the Amazon Alexa dataset in Section~\ref{label:experiments}. Finally, we conclude the paper in Section~\ref{sec:conclusion}.
\section{Motivations}
\label{sec:motivation}
\begin{figure*}[!ht]
	\centering
	\includegraphics[width=0.95\textwidth]{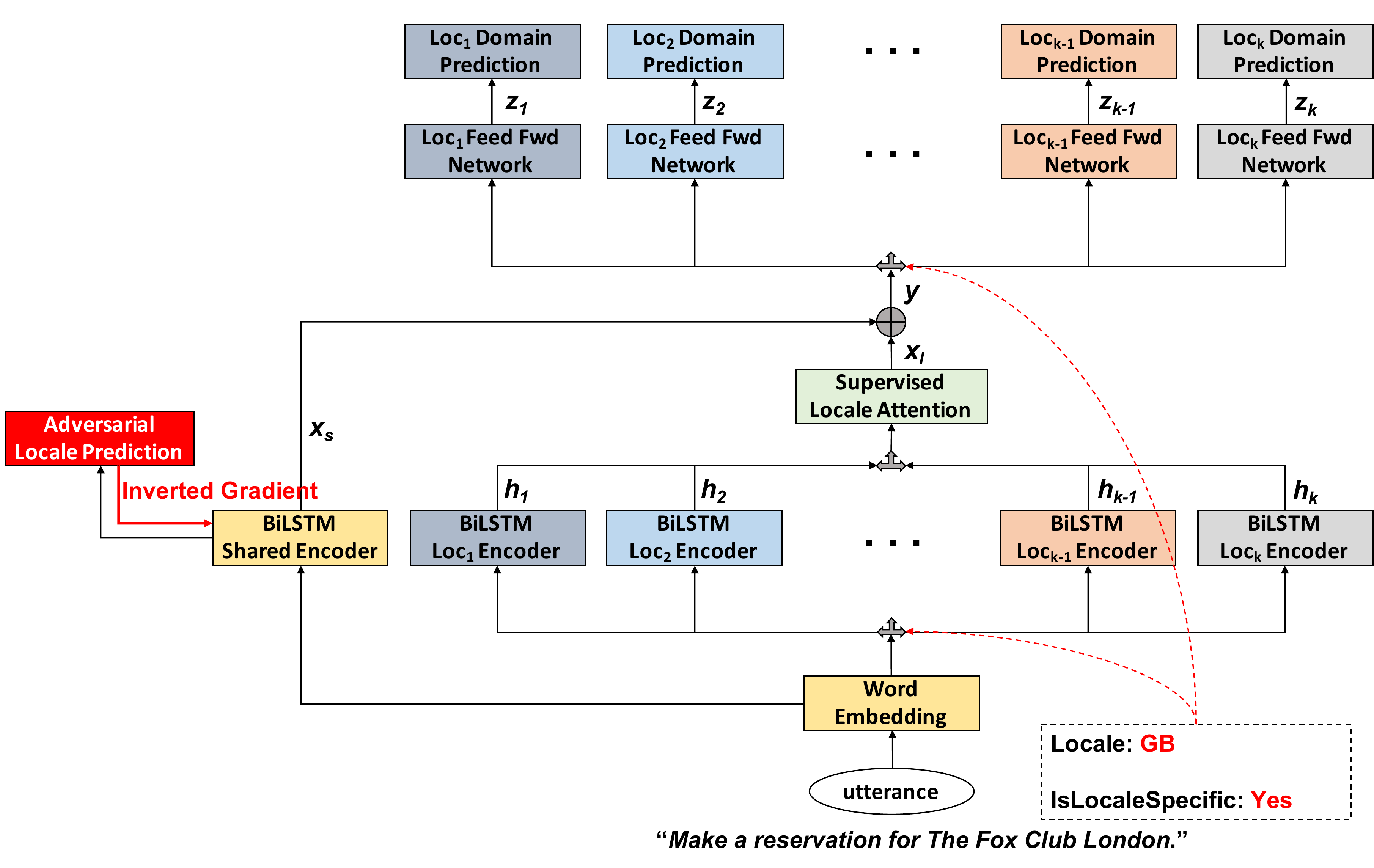}
	\caption{Model architecture of the universal model.}
	\label{fig:model_architecture}
\end{figure*}


\subsection{Locale/Domain-Maturity}
Let the term \textit{maturity} be defined by how long it has been since a service or model was deployed in a locale and/or how much data have been collected. Every locale has different degrees of maturity. That is, while some locales have spent time long enough to collect sufficient data to train models, others may suffer from the lack of data (see more details of data statistics in Section~\ref{label:experiments}). In addition to that, domains that are commonly available in multiple locales have different levels of maturity for each locale. Those two dimensions of maturity are not always aligned with each other. In other words, there could exist domains that have more data in immature locales than in mature locales, depending on targeted users, regional properties of domains, and so forth.

\subsection{Locale-Specificity}
When an SLU service is deployed in multiple locales, each of the locales has its own domain set and there can exist overlapping domains between locales. Such domains may share the same schema including intents and slots and thus they should be able to handle the same patterns of utterances regardless of locales. It allows locales to share the knowledge of common domains with each other, which eventually helps immature locales to overcome the lack of data. A special case that needs to be carefully considered is that a domain could be locale-specific. Even though a domain is common across different locales, it may be defined with different intents/slots. For example, the domain \texttt{OpenTable}, which is capable of restaurant reservation, is available in both US and GB, but the slot values including restaurant names are totally different between the two locales. That is, the utterance \textit{``Make a reservation for The Fox Club London''} can be handled by \texttt{OpenTable} in GB locale, but probably not in US locale, because the restaurant \textit{The Fox Club London} is located in London. If we have different locales share the same utterance patterns between them even for such locale-specific domains, then it will cause confusion on the models. We thus identify locale-specific domains in advance of model training and do not allow input utterances of such domain to be shared by different locales. We need to handle domains in a similar way that are available only in a particular locale.
\section{Universal Model}
\label{label:universal_model}
In this section, we describe our proposed model illustrated in Figure~\ref{fig:model_architecture} in detail. Suppose that given $k$ locales, $\{l_i | i=1,2,\dots,k\}$, each locale $l_i$ is associated with its own domain set $D_i = \{d_{ij} | j = 1, 2, \dots, |D_i| \}$. There could exist overlapping domains between locales and some of the overlapping domains may share exactly the same intents/slots while others may have different intents/slots across locales. The main task is that given an input utterance from locale $l_i$ the model should be able to correctly classify the utterance into a domain $d_{ij} \in D_i$ that can best handle the utterance. Here we assume that all locales use the same language, English, but have different domain sets. Our deep neural model, as a proposed solution to the task, is comprised of two layers. The first layer includes a BiLSTM shared encoder and $k$ BiLSTM locale-specific encoders. The second layer consists of a set of $k$ locale-specific prediction layers.

\subsection{Shared and Locale-specific Encoders}
Given an input utterance that forms a word sequence, an encoder makes a vector representation of the entire utterance by using word embeddings for English language in general. We use Bidirectional LSTM (BiLSTM) to encode an input utterance and consider it to be a mapping function $\mathcal{F}$ that consumes a sequence of word embeddings and then produces an embedding vector given by concatenating the outputs of the ends of the word sequences from the forward LSTM and the backward LSTM. While different locales share common domains and utterances, each of them also should be able to learn certain patterns observed from domains available only in the locale. In other words, there exist both global and local patterns in the entire domain set. In order to effectively capture both patterns and avoid confusion between locales, we use a shared encoder $\mathcal{F}_s$ and multiple locale-specific encoders $\mathcal{F}_{l_i}$ for $\forall i=1, 2, \dots, k$, each of which corresponds to a particular locale $l_i$, as similarly adopted in~\citep{kim2017cross,kim2016frustratingly}. While the shared encoder $\mathcal{F}_s$ learns global patterns of utterances commonly observable across different locales, each of the locale-specific encoders $\mathcal{F}_{l_i}$, which corresponds to one of the locales $l_i$, learns local patterns of utterances that are observed specifically in the locale $l_i$.

\subsection{Adversarial Locale Prediction Loss}
\label{sec:adversarial_loss}
Intuitively, the shared encoder $\mathcal{F}_{s}$ is expected to be able to better capture common utterance patterns over all locales rather than to learn patterns that are seen in only some particular locales. Thus, $\mathcal{F}_s$ can be further tuned to be locale-invariant by adding a locale prediction layer with negative gradient flow, as similarly proposed in~\citep{kim2017adversarial,ganin2016domain,Liu2017AdversarialML}. Let $\mathbf{x_s}$ denote an encoded vector for an input utterance produced by the shared encoder $\mathcal{F}_s$. $\mathbf{x}_s$ is then fed into a single-layer neural network to make a prediction for its corresponding locale $l_i$. Formally,

\begin{equation}
    \mathbf{z}_{adv} = softmax(\mathbf{W}_{adv} \cdot \mathbf{x}_s + \mathbf{b}_{adv})
\end{equation}
where $\mathbf{W}_{adv}$ and $\mathbf{b}_{adv}$ are a weight matrix and a bias term for the locale prediction layer of the feed-forward network. Since we aim to make the shared encoder $\mathcal{F}_s$ to be locale-invariant, the adversarial locale prediction loss is given by the \textit{positive} log-likelihood:

\begin{equation}
    \mathcal{L}_{adv} = \displaystyle \sum_{i=1}^{k} t_i \log [\mathbf{z}_{adv}]^{i}
\end{equation}
where $t_i$ is a binary indicator if locale $l_i$ is the correct prediction or not.

\subsection{Supervised Locale Attention}
In order to allow the locale-specific encoders to share knowledge about common domains across locales, we give a chance to learn an input utterance to any locale-specific encoders $\mathcal{F}_{l_i}$ as long as its associated domain is in $D_i$, except the case of locale-specific domains (i.e., \texttt{OpenTable}).
Suppose $S_{d_{ij}}= \{l_w | d_{ij} \in D_w, \forall w = 1, 2, \cdots, k\}$ if $d_{ij}$ is not locale-specific, otherwise $S_{d_{ij}} = \{l_i\}$.
That is, depending on which locales a given domain is available in and whether or not it is locale-specific, its utterance needs to be selectively routed to locale-specific encoders $\mathcal{F}_{l_i}$ where $l_i \in S_{d_{ij}}$. However, we do not know a ground-truth domain associated with an input utterance during inference and it means that there is no way to do such selective routing unfortunately. Instead, we can use supervised attention mechanism to approximate the locales in which a domain is available. During training, we have each of the locale-specific encoder outputs attend each other and provide them with information about which locales should be highly attended, as explained in the following.

Let $\mathbf{H} = [\mathbf{h}_{l_1}, \mathbf{h}_{l_2}, \dots, \mathbf{h}_{l_k}] \in \mathds{R}^{d_h \times k}$ denote a matrix of encoded vectors generated by $\mathcal{F}_{l_i}$ for $\forall i = 1, 2, \dots, k$. Then, the attention weights are obtained as follows,

\begin{equation}
	\mathbf{a} = logistic(\mathbf{w} \cdot tanh(\mathbf{V} \cdot \mathbf{H}))
\end{equation}
where $\mathbf{w} \in \mathds{R}^{d_a}$ and $\mathbf{V} \in \mathds{R}^{d_a \times d_h}$ are learnable weight parameters, and $d_a$ is a hyperparameter we can set arbitraily. The resulted vector $\mathbf{a}$ contains attention weights in the range between 0 and 1 over the encoded vectors $\mathbf{h_{l_1}}, \dots, \mathbf{h_{l_k}}$. Then a locale-aware encoded vector $\mathbf{x_l}$ can be achieved by taking a weighted linear combination of $\mathbf{h_{l_1}}, \dots, \mathbf{h_{l_k}}$:

\begin{equation}
	\mathbf{x}_l = \mathbf{a} \cdot \mathbf{H}^{\top}
\end{equation}

The final vector representation $\mathbf{y} \in \mathds{R}^{2 \cdot d_h}$ for the input utterance is the concatenation of two encoded vectors $\mathbf{x}_s$ and $\mathbf{x}_l$ that are produced from $\mathcal{F}_s$ and $\mathcal{F}_{l}$, respectively. Note we have to make sure that the proper encoders that correspond to $S_{d_{ij}}$ always get high attention weights. Thus, instead of just letting $\mathbf{V}$ and $\mathbf{w}$ be optimized during training the model, we can optimize them in a supervised way. That is, in training time, the model is aware of locales where a ground-truth domain is available. In other words, we can reward or penalize the attention weights depending on whether or not their corresponding locales have the domain of an input utterance. Therefore, the loss function for the attention weights is defined as,

\begin{equation}
	\mathcal{L}_{loc} = -\big( \displaystyle \sum_{l \in S_{d_{ij}}} \log(a_l) + \sum_{l' \notin S_{d_{ij}}} \log(1 - a_{l'}) \big)
\end{equation}

\subsection{Domain Classification}
Once we obtain an encoded vector $\mathbf{y}$ that represents an input utterance, we feed it into prediction layers, consisting of feed forward networks, to make predictions. Since the availability of domains depends on locales, the prediction layers use the locale information associated with the utterance to route the encoded vector to only a subset of prediction layers in which the domain of the utterance is available. Then, the output vector produced by the prediction layer specifically for the locale $l_i$ is

\begin{equation}
	\mathbf{z}_{i} = \mathbf{W}_{i}^2 \cdot \sigma(\mathbf{W}_{i}^1 \cdot \mathbf{y} + \mathbf{b}_{i}^1) + \mathbf{b}_{i}^2
\end{equation}
where $\mathbf{W}_{i}$ and $\mathbf{b}_{i}$ are the weight and bias parameters used by the $l_i$ specific prediction layer, and $\sigma$ is an activation function for non-linearity. Since our model is structured with a multi-task learning framework to learn a joint representation across locales, we calculate $\mathbf{z}_{i}$ for all $l_i \in S_{d_{ij}}$ and then the predictions are made independently. Then the prediction loss is

\begin{align}
	\mathcal{L}_{pos} & = - \log p(d_{ij} | z_i) \\
	\mathcal{L}_{neg} & = - \sum_{\substack{\hat{d_{ij}} \in D_i \\ \hat{d_{ij}} \neq d_{ij}}} \log p(\hat{d_{ij}} | z_i) \\
	\mathcal{L}_{pred} & = \displaystyle \frac{1}{|S_{d_{ij}}|}\sum_{l_i \in S_{d_{ij}}} (\mathcal{L}_{pos} + \mathcal{L}_{neg})
\end{align}
Note that the prediction loss must be normalized by the number of locales in $S_{d_{}ij}$ because the size of the set changes depending on how many locales has the domain associated with an input utterance and thus the number of the final prediction layer
Then, the final objective function looks as follows,

\begin{equation}
	\argmin_{\theta_{\mathcal{F}_s}, \theta_{\mathcal{F}_{l}}, \mathbf{V}, \mathbf{w}, \mathbf{W}, \mathbf{b}} \mathcal{L}_{adv} + \mathcal{L}_{loc} + \mathcal{L}_{pred}
\end{equation}
where $\theta_{\mathcal{F}_s}$ and $\theta_{\mathcal{F}_l}$ are the LSTM weight parameters in the shared encoder and the locale-specific encoders, respectively.
\section{Experiments}
\label{label:experiments}
\subsection{Dataset}
\label{sec:dataset}
We use a subset of the Amazon Alexa dataset that consists of utterances spoken to Alexa by real customers over four different English locales including US (United States), GB (United Kingdom), CA (Canda), IN (India). Each of the utterances is labeled with a ground-truth domain. The main objective of this experiment should be to show the effectiveness of various approaches on domain classification task under the situation where there exist multiple locales that have imbalanced data and disparate domain sets. Thus, we consider the following two aspects: 1) how differently various domain classification approaches behave depending on domains and 2) how well they can overcome the challenging issues discussed in Section~\ref{label:introduction}. To this end, we categorize all domains in the dataset into four different groups.

\begin{itemize}
    \item \textbf{Locale-specific} A set of domains which are defined with different intents/slots across locales.
    \item \textbf{Locale-independent} A set of domains which have exactly the same intent/slot lists across locales.
    \item \textbf{Single-locale} A set of domains which are available in only a single locale.
    \item \textbf{Small} A set of domains that lack data in a locale but have sufficient data in other locales.
\end{itemize}

Table~\ref{tab:data} shows its brief statistics per locale, Table~\ref{tab:testset} presents the number of domains for each of four different domain categories, and Table~\ref{tab:domain_overlap} shows how many domains are overlapping between locales.

\begin{table}[!t]
	\centering
	\scriptsize
	\begin{tabular}{| c | c | c | c | c |}
		\hline
		\textbf{Locale} & \textbf{Train} & \textbf{Validation} & \textbf{Test} & \textbf{No. domains} \\ \hline
		US & 173,258 & 24,653 & 122,931 & 177 \\ \hline
		GB & 85,539 & 10,378 & 53,226 &  240 \\ \hline
		CA & 7,113 & 887 & 4,487 & 51 \\ \hline
		IN & 4,821 & 637 & 2,990  & 41 \\ \hline
	\end{tabular}
	\caption{Data statistics}
   	\label{tab:data}
\end{table}

\begin{table}[!t]
	\centering
	\scriptsize
	\begin{tabular}{| c | c | >{\centering\arraybackslash}p{0.7cm} | >{\centering\arraybackslash}p{1.2cm} | >{\centering\arraybackslash}p{1.1cm} | c |}
		\hline
		\textbf{Locale} & \textbf{Overall} & \textbf{Locale-specific} & \textbf{Locale-independent} & \textbf{Single-locale} & \textbf{Small} \\ \hline
		US & 177 & 15 & 162 & 0 & 35 \\ \hline
		GB & 240 & 16 & 224 & 82 & 100 \\ \hline
		CA & 51 & 3 & 48 & 6 & 33 \\ \hline		
		IN & 41 & 4 & 37 & 12 & 20 \\ \hline
	\end{tabular}	
	\caption{Test set breakdown}
    \label{tab:testset}
\end{table}

\begin{table}[!t]
	\centering
	\scriptsize	
	\begin{tabular}{| c | c | c | c | c |}
		\hline
		& \textbf{US} & \textbf{GB} & \textbf{CA} & \textbf{IN} \\ \hline
		US & 177 & 155 & 44 & 26 \\ \hline
		GB &  & 240 & 27 & 23 \\ \hline
		CA & & & 51 & 10 \\ \hline		
		IN & & & & 41  \\ \hline
	\end{tabular}
	\caption{Domain overlaps between locales}
    \label{tab:domain_overlap}
\end{table}

\begin{table*}[!t]
	\centering
	\small
	\begin{tabular}{| c | c | c | c | c | c | c |}
		\hline
		\textbf{Locale} & \textbf{Model} & \textbf{Overall} & \textbf{Locale-specific} & \textbf{Locale-independent} & \textbf{Single-locale} & \textbf{Small} \\ \hline
		\multirow{5}{*}{US}
		& single & 70.21 & 54.39 & 69.90 & -- & 8.18 \\
		& union & 70.21 & 54.39 & 69.90 & -- & 8.18 \\
		& constrained & 74.25 & 76.08 & 74.02 & -- & 38.30 \\
		& universal & \textbf{82.64} & 88.20 & \textbf{81.92} & -- & \textbf{61.79} \\
        & universal + adv & 11.13 & \textbf{97.51} & 0.00 & -- & 5.38 \\ \hline
		\multirow{5}{*}{GB}
		& single & 56.02 & 62.81 & 55.09 & 37.81 & 0.00 \\
		& union & 66.61 & 78.74 & 64.96 & 48.19 & 36.54 \\
		& constrained & 67.82 & 76.83 & 66.60 & 50.51 & 38.04 \\
		& universal & 80.06 & \textbf{88.37} & 78.93 & \textbf{83.60} & 57.96 \\
        & universal + adv & \textbf{80.52} & 85.88 & \textbf{79.79} & 82.22 & \textbf{59.52} \\ \hline
		\multirow{5}{*}{CA}
		& single & 43.43 & 3.57 & 43.68 & 0.00 & 0.24 \\
		& union & 61.04 & 10.71 & 61.35 & 0.65 & 30.78 \\
		& constrained & 76.46 & 67.85 & 76.51 & 39.17 & 55.66 \\
		& universal & \textbf{94.00} & \textbf{75.00} & \textbf{94.12} & 97.74 & \textbf{77.09} \\
        & universal + adv & 35.21 & 71.42 & 34.98 & \textbf{98.87} & 36.69 \\ \hline
		\multirow{5}{*}{IN}
		& single & 56.25 & 0.00 & 60.46 & 0.00 & 0.00 \\
		& union & 45.93 & 0.00 & 49.38 & 0.00 & 17.96 \\
		& constrained & 62.64 & 44.71 & 63.98 & 25.94 & 58.64 \\
		& universal & \textbf{88.09} & \textbf{87.01} & \textbf{88.17} & 80.00 & \textbf{68.47} \\
        & universal + adv & 22.30 & \textbf{87.01} & 17.46 & \textbf{82.97} & 10.50 \\ \hline
	\end{tabular}
	\caption{Domain classification accuracy over different domain categories and different locales.}
    \label{tab:domain_classification}
\end{table*}

\subsection{Competing Models}
We compare the performances of the following five models.

\begin{itemize}
    \item \textbf{single} A standard BiLSTM based encoder trained with only data in a particular locale.
    \item \textbf{union} An extension of `single' trained with US data additionally.
    \item \textbf{constrained} A BiLSTM encoder trained with all locales data. It uses the locale information associated with the utterance to route the encoded utterance to only a subset of domains available in the constrained output space for the locale to make prediction~\citep{kim2016domainless,kim2016natural}.
    \item \textbf{universal} This is our main contribution model described throughout the paper.
    \item \textbf{universal + adv} An extension of `universal' incorporating the adversarial locale prediction loss as discussed in Section~\ref{sec:adversarial_loss}.
\end{itemize}


\subsection{Domain Classification}
To demonstrate the effectiveness of our model architecture especially on domains with insufficient data and/or locale-dependency, we report the classification performances of all competing models on several subsets of the dataset (four different groups presented in Section~\ref{sec:dataset}) as well as the entire data. We use classification accuracy as our main evaluation metric. The experimental results in Table~\ref{tab:domain_classification} clearly show two major points: 1) our proposed universal model outperforms all other baselines over all locales and all domain sets, and 2) the baseline models achieve very poor accuracy especially when leveraging available data in other locales is of critical importance or when there needs to selectively share knowledge depending on the locale-specificity of a domain. If a model that shares knowledge across locales does not handle locale-specific domains carefully, its performance would deteriorate due to confusion on locale-specific patterns. The `constrained' model uses a shared encoder and allows locales to shares its prediction layer, but it does not determine whether or not to share knowledge for each domain. As a result, its classification accuracy is only 44\% for locale-specific domains and 25\% for single-locale domains in the IN dataset with lack of data. Also, `single' and `union' models do not have any chance to learn a joint representation while sharing knowledge and thus they totally fail to make predictions correctly for locale-specific, single-locale, and small domains. In contrast, our universal model is very robust to domains with insufficient data and domains with locale-specific patterns over all locales. It proves that our approach is very effective for capturing both global and local patterns by selectively sharing domain knowledge across locales. Also, the adversarial locale prediction is only helpful for locale-specific and single-locale domains. That is probably because the effect of adversarial loss paradoxically makes the model rely on only the locale-specific encoders which are well-optimized for locale-specific/single-locale domains. There needs deep analysis about why it does not affect the GB locale, and we leave it as future works.

\subsection{Implementation Details}
All the models were optimized using a minibatch size of 64 and trained for 20 epochs by the Adam optimizer~\citep{kingma2014adam} with initial parameter values $\eta = 1 \times 10^{-3}, \beta_{1} = 0.9, \beta_{2} = 0.999$. We picked the weight parameter values that achieved the best classification accuracy on the validation set to report the test set accuracy presented in Table~\ref{tab:domain_classification}. We used pre-trained word embeddings with 100 dimensionality, generated by GloVe~\citep{pennington2014glove}. The dimensionality of each hidden output of LSTMs is 100 for both the shared encoder $\mathcal{F}_s$ and the locale-specific encoder $\mathcal{F}_{l_i}$, and the hidden outputs of both forward LSTM and backward LSTM are concatenated, thereby the output of each BLSTM for each time step is 200. The inputs and the outputs of the
BLSTMs are regularized with dropout rate 0.5~\citep{pham2014dropout}.
\section{Conclusion}
\label{sec:conclusion}
In this paper, we propose a multi-task learning based locale-agnostic universal model for domain classification task that dynamically chooses subsets of locale-specific components depending on input data. It leverages available data across locales sharing the same language to reduce the cost of scaling the domain classification model to a larger number of locales and maximize model performance even for new locales without sufficient data. The experimental results show that the universal model effectively exploits both global and local patterns and allows locales selectively share knowledge with each other.
Especially, its classification performance is notable on immature locales/domains with insufficient data and locale-specific domains.

For future work, we consider adopting the proposed model architecture to multi-lingual scenario as well. The proposed model architecture is limited to supporting multiple locales using the same language only (e.g.., English in our experiments). However, voice-driven virtual assistant systems are becoming more and more popular around the world while expanding to non-English locales such as France, Italy, Spain and so on, and there could be a lot of domains built with multiple supported languages. It will definitely make the rapid scaling of a domain classification model to a large number of locales much more challenging in the future. We plan to address several issues, including but not limited to: 1) how can we capture and share knowledge of common patterns of utterances belonging to the same domain but written in different languages across different locales? 2) how can we prevent a locale from interfering with other locales using different language for learning linguistic context of utterances?

\bibliography{naaclhlt2019}
\bibliographystyle{acl_natbib}

\end{document}